\documentclass{article}

\usepackage[final, nonatbib]{midl_2018}

\usepackage[utf8]{inputenc} 
\usepackage[T1]{fontenc}    
\usepackage{hyperref}       
\usepackage{url}            
\usepackage{booktabs}       
\usepackage{amsfonts}       
\usepackage{nicefrac}       
\usepackage{microtype}      
\usepackage{graphicx}
\usepackage{subfigure}

\title{Contextual Hourglass Networks for Segmentation and Density Estimation}

\author{
  Daniel O\~noro-Rubio \\
  NEC Labs Europe\\
  \texttt{daniel.onoro@neclab.eu} \\
  \And
  Mathias Niepert \\
  NEC Labs Europe\\
  \texttt{mathias.niepert@neclab.eu} \\
}

\begin{document}

\maketitle

\section{Introduction}

Hourglass networks such as the U-Net~\cite{unetRonnebergerFB15} and V-Net~\cite{MilletariNA16} are popular neural architectures for medical image segmentation and counting problems. Typical instances of hourglass networks contain shortcut connections between mirroring layers. These shortcut connections improve the performance and it is hypothesized that this is due to mitigating effects on the vanishing gradient problem and the ability of the model to combine feature maps from earlier and later layers. We propose a method for not only combining feature maps of mirroring layers but also feature maps of layers with different spatial dimensions. For instance, the method enables the integration of the bottleneck feature map with those of the reconstruction layers. The proposed approach is applicable to any hourglass architecture. We evaluated the contextual hourglass networks on image segmentation and object counting problems in the medical domain. We achieve competitive results outperforming popular hourglass networks by up to 17 percentage points.

%

\section{Contextual Convolutions}

\begin{figure}
\centering
\subfigure[Simple U-Net with contextual convolutions.] {
  \includegraphics[width=0.55\linewidth]{contextual_hourglass_diagram_new.pdf}
  \label{fig:context_diagram}
}\hspace{8mm}
\subfigure[Contextual convolution.] {
  \includegraphics[width=0.34\linewidth]{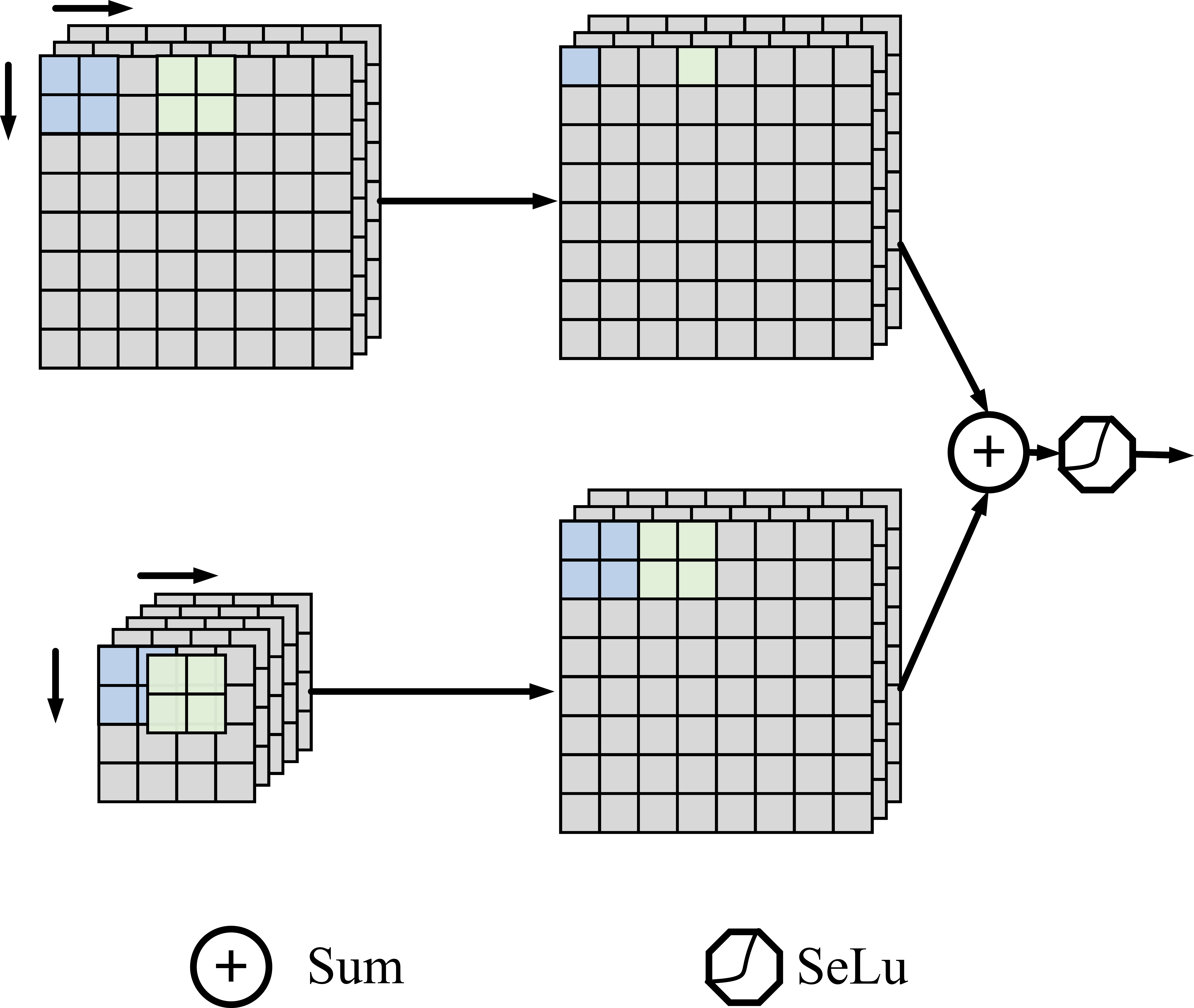}
  \label{fig:context_convolution}
}
\caption{An illustration of the proposed contextual hourglass networks where shortcut connections between the bottleneck and later representations are established with the contextual convolution operation that enforces a spatial alignment of the resulting feature maps. }
\label{fig:general_diagram}
\end{figure}

Intuitively, hourglass networks have two stages. In the first stage, an image is encoded with each convolutional layer into a more compressed and spatially smaller representation. During this encoding process, the scope of the receptive field increases. We refer to the most compressed representation located at the center of the network as the bottleneck representation. In the second stage, every transpose convolutional layer decodes an increasingly less compressed and spatially larger representation beginning with the bottleneck. This is often referred to as the decoding stage. 
In several hourglass type networks such as the U-Net, every layer from the second stage is connected to its mirroring layer in the first stage. These shortcut connections perform some aggregation operation such as a summation or concatenation between the respective feature maps. Figure \ref{fig:context_diagram} illustrates a simple hourglass architecture with two layers in the first stage, a bottleneck representation, and two layers in the second stage. The shortcut connections between mirroring layers are indicated with dashed lines. The aggregation operation is a concatenation.  

In the proposed contextual hourglass networks, there are additional shortcut connections between layers of differing spatial dimensions. This has several advantages. First, it allows to incorporate the bottleneck representation in later spatially more extensive layers -- the bottleneck representation provides a \emph{context} for the decoding layers. Second, it facilitates a more direct flow of gradients from the output layer to the more compressed representations such as the bottleneck. 
The main contribution of this paper is a mechanism to spatially tie two different filter banks and their movement over two feature maps of differing size. Let $\mathbf{T}_1$ be a feature map of dimension $w_1 \times h_1 \times d_1$, that is, a feature map with width $w_1$, height $h_1$ and with $d_1$ channels. Moreover, let $\mathbf{T}_2$ be a feature map of dimension $w_2 \times h_2 \times d_2$ with $w_2 > w_1$ and $h_2 > h_1$. Here, $\mathbf{T}_1$ is a more compressed feature map of an earlier layer. $\mathbf{T}_2$ is a less compressed feature map, with larger spatial extent, and the output of a later layer in an hourglass type network. To create a shortcut connection between the respective layers and to apply an aggregation function between spatially aligned feature maps, we tie the movement of the filter bank of the convolutional layer operating on $\mathbf{T}_1$ to the movement of the filter bank operating on $\mathbf{T}_2$. Under the assumption that both convolutional layers have the same stride of $1$ and the same padding strategy, one movement of the filter bank on $\mathbf{T}_2$ to the right (to the bottom) corresponds to $\left\lfloor w_1 / w_2 \right\rfloor$ movements to the right ($\left\lfloor h_1 / h_2 \right\rfloor$ to the bottom) of the filter bank on $\mathbf{T}_1$.
Figure \ref{fig:context_diagram} illustrates the addition of contextual convolutions to an hourglass type network. For the sake of simplicity, we only depict the height and depth of the layers' feature maps. A contextual convolution connects the bottleneck representations with later layers in the network. The crucial property of the contextual convolutions is the spatial alignment between the resulting feature maps. Figure \ref{fig:context_convolution} illustrates the entanglement of movements of the filter bank on two tensors of different sizes. A movement on the spatielly more extensive tensor corresponds to a fraction of a movement on the smaller tensor. We obtained good results with a summation operation and the SeLu activation function. Note that contectual convolutions can be added to hourglass architectures such as the U-Net~\cite{unetRonnebergerFB15}, V-Net~\cite{MilletariNA16}, and the Tiramisu net~\cite{JegouDVRB17}.

\section{Experimental Evaluation}

\subsection{Image Segmentation}

We perform experiments on the EM segmentation challenge data set of ISBI 2012. The dataset is composed of 60 grayscale images of $512 \times 512$ pixels. There are 30 labeled and 30 unlabeled images. We trained the networks parameters by randomly sampling 25 labeled images for training and 5 for validating. Finally, we segmented the 30 unlabeled test images and obtained the resulst by sending those to the organizers of the challenge. We compare the proposed contextual hourglass networks with the Tiramisu and U-Net. To ensure a fair comparison we also use the SeLu activation and the exact same number of layers for the U-Net. We refer to the contextual hourglass architecture as the ``Contextual U-Net.''
All models are trained under the same conditions. We randomly initialize all the weights with the Xavier method \cite{xavier2010}, alternatively, and with a similar performance, we have also tried He-Uniform \cite{He_2015_ICCV}. During the training we optimize the categorical cross entropy loss. The training strategy consists of two parts. In the first part, we train on augmented the data by performing randomly distortions. In the second step, we fine tune the models for the nondistorted data. On each part the models are trained until convergence.
Table \ref{tab:isbi_results} lists the results. The contextual U-Net significantly outperforms the other networks showing that the contextual convolutions lead to a significant improvement over the U-Net architecture. Figure \ref{fig:qualitatives_isbi} depicts some qualitative result on the validation set.

\begin{table}[h]
\small
  \caption{ISBI results on the test set.}
  \label{tab:isbi_results}
  \centering
  \begin{tabular}{lcc}
	\toprule
    Method					& Rand Score Thin	& Information Score Thin \\
    \midrule
    Tiramisu-103 \cite{JegouDVRB17}	& 0.7628			& 0.9165     \\
    U-Net \cite{unetRonnebergerFB15} 	& 0.8737			& 0.9594     \\
    Contextual U-Net					& 0.9366			& 0.9737     \\
    \bottomrule
  \end{tabular}
\end{table}

\subsection{Object Counting}

\begin{figure}
\centering
\subfigure[ISBI data set] {
  \includegraphics[width=0.619\linewidth]{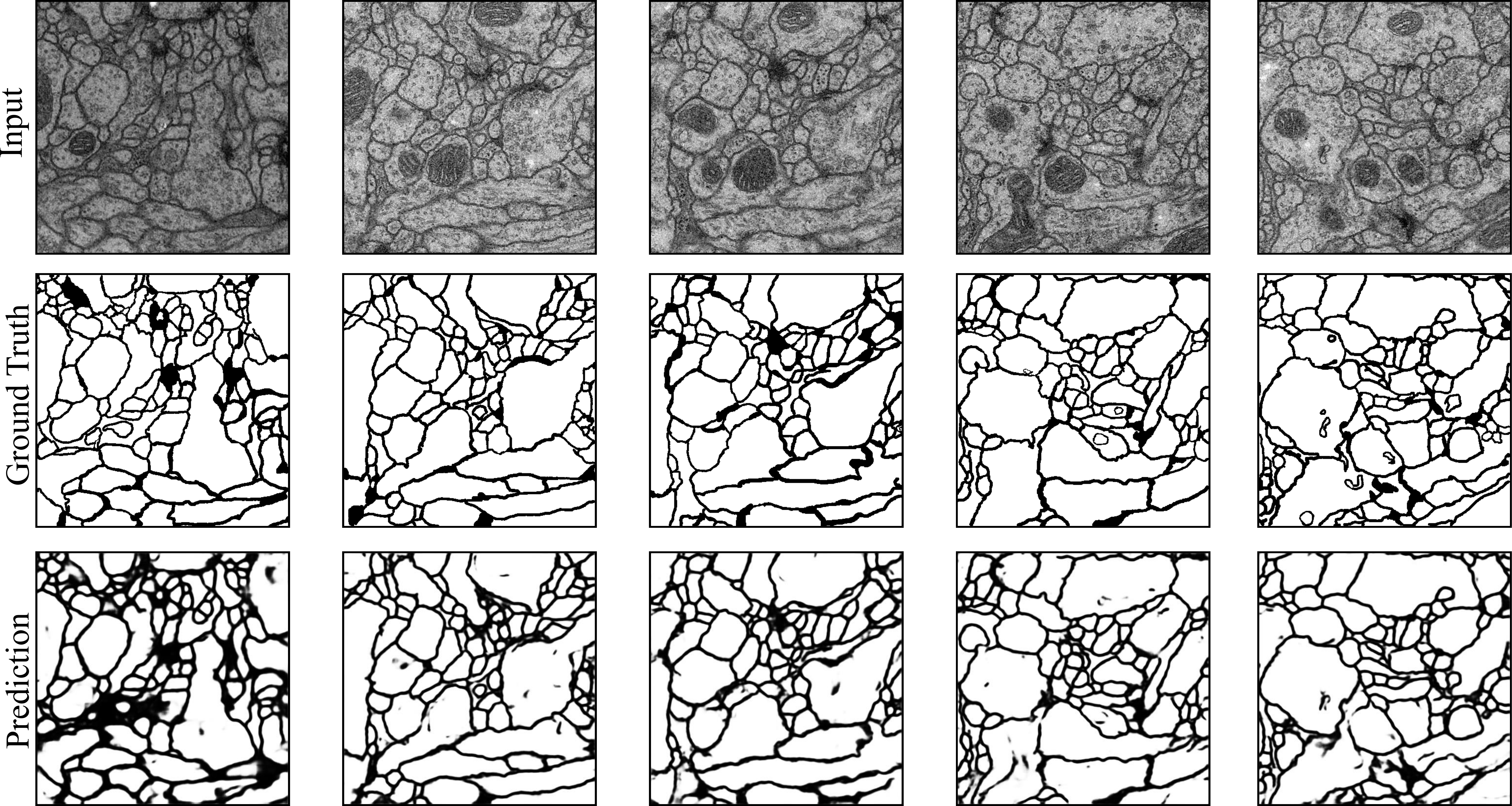}
  \label{fig:qualitatives_isbi}
}
\subfigure[Cell counting data set] {
  \includegraphics[width=0.347\linewidth]{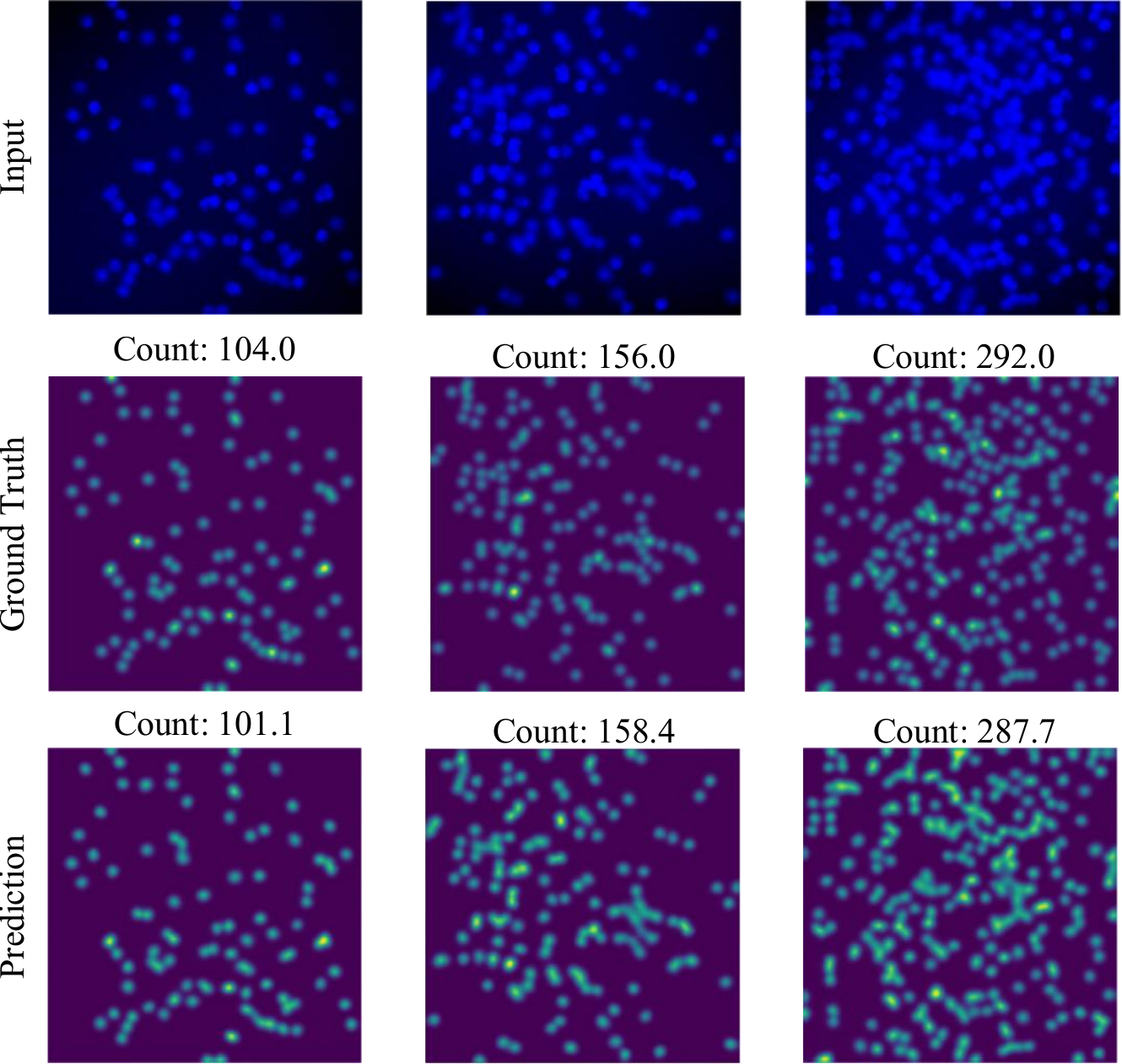}
  \label{fig:qualitatives_cells}
}
\caption{Qualitative results for the datasets. Predictions are generated by the contextual U-net.}
\end{figure}

We apply the proposed model class to the different problem of cell counting. For this task, we use the simulated fluorescence microscope images of \cite{LehmussolaRSHY07}. We followed the exact same experimental setup as in previous work~\cite{lempitsky2010}. The dataset consists of 200 images. We used the first 32 images for training, the 68 following images for validation, and the last 100 images for testing. We used a simplified variant of the contextual U-Net with 3 encoding and 3 decoding steps and set the number of base filters to 24. We used the mean squared difference as loss function. We train our model from scratch by initialing its weights with the Xavier algorithm. We perform data augmentations such as random perturbations and the network is trained until  convergence.
Table \ref{tab:counting-results} shows the mean absolute error for N training images. Despite the reduced amount of training data, the proposed model achieve a competitive performance compared with the current state-of-the-art. In Figure \ref{fig:qualitatives_cells} we present some qualitative results.

\begin{table}[h]
\small
  \caption{Cell counting results.}
  \label{tab:counting-results}
  \centering
  \begin{tabular}{lcccccc}
	\toprule
    Method					& N = 1	& N = 2 & N = 4 & N = 8 & N = 16 & N = 32\\
    \midrule
    Linear regression \cite{lempitsky2010}	& 67.3 $\pm$ 25.2 & 37.7 $\pm$ 14.0 & 16.7 $\pm$ 3.1 & 8.8 $\pm$ 1.5 & 6.4 $\pm$ 0.7 & 5.9 $\pm$ 0.5 \\
    Detection \cite{lempitsky2010}	& 28.0 $\pm$ 20.6 & 20.8 $\pm$ 5.8 & 13.6 $\pm$ 1.5 & 10.2 $\pm$ 1.9 & 10.4 $\pm$ 1.2 & 8.5 $\pm$ 0.5 \\    
    MESA \cite{lempitsky2010}	& 9.5 $\pm$ 6.1 & \textbf{6.3} $\pm$ 1.2 & 4.9 $\pm$ 0.6 & 4.9 $\pm$ 0.7 & \textbf{3.8} $\pm$ 0.2 & \textbf{3.5} $\pm$ 0.2 \\
    Contextual U-Net		& \textbf{6.5} & 6.5 & \textbf{3.8} & \textbf{3.8} & 5.2 & 4.4 \\
    \bottomrule
  \end{tabular}
\end{table}

\bibliographystyle{abbrv}
\bibliography{references}

\end{document}